\pdfoutput=1

\documentclass[11pt]{article}

\usepackage[preprint]{acl}

\usepackage{times}
\usepackage{latexsym}
\usepackage[T1]{fontenc}

\usepackage[utf8]{inputenc}

\usepackage{microtype}

\usepackage{inconsolata}

\usepackage{graphicx}

\usepackage{tikz}
\usepackage{pgf-pie}
\usepackage{amsmath}
\usepackage{multirow}
\usepackage{booktabs}
\usepackage{array}
\usepackage{makecell}
\usepackage{tabularx}
\usepackage{textcomp,mathcomp}
\newcolumntype{H}{>{\setbox0=\hbox\bgroup}c<{\egroup}@{}}
\usepackage{shortcuts}
\usepackage{comment}
\usepackage{footmisc}

\usepackage{url}

\usepackage{enumitem}	
\setlist{leftmargin=1em}
\usepackage{setspace}

\AtBeginDocument{%
  }

\def\Snospace~{\S{}}

\title{Detecting Manipulated Contents Using Knowledge-Grounded Inference} 

\author{Mark Huasong Meng \\
	Technical University of Munich \\
	\\\And
	Ruizhe Wang, Meng Xu\\
	University of Waterloo \\\And
	Chuan Yan, Guangdong Bai\\
	The University of Queensland\\
	}

\begin{document}
 \maketitle
\begin{abstract}

The detection of manipulated content, a prevalent form of fake news, has been widely studied in recent years. 
While existing solutions have been proven effective in fact-checking and analyzing fake news based on historical events, 
the reliance on either intrinsic knowledge obtained during training or manually curated context hinders them from tackling zero-day manipulated content which can only be recognized with real-time contextual information. 

In this work, we propose \toolname, a tool designed for detecting zero-day manipulated content. 
\toolname first sources contextual information about the input claim from mainstream search engines, and subsequently vectorizes the context for the large language model (LLM) through retrieval-augmented generation (RAG). 
The LLM-based inference can produce a ``truthful'' or ``manipulated'' decision and offer a textual explanation for the decision.
To validate the effectiveness of \toolname, we also propose a dataset comprising \allfake pieces of manipulated fake news derived from \allnewscollected recent real-world news headlines. 
\toolname achieves an overall F1 score of \ourfone on this dataset and outperforms existing methods by up to 1.9x in F1 score on their benchmarks on fact-checking and claim verification.

\textcolor{red}{Warning: This paper contains manipulated content that may be offensive, harmful, or biased.}
\end{abstract}


\section{Introduction}
\label{sec:intro}

The widespread mis- and disinformation
(colloquially known as ``fake news''~\cite{ITSAP.00.300})
has become a long-lasting cyber threat in the Internet era~\cite{web:gallup2020fake,zurko2022disinformation}.
In particular,
disinformation
%
caused by manipulated content has been ranked among
the biggest sources of cybercrime and
the most potent cybersecurity threats~\cite{mazurczyk2024disinformation}.
Manipulated content,
as the name suggests,
is typically produced based on true information
(e.g., events that happened in the real world)
but contains subtle fabrications for malicious purposes.
Without proper detection and filtering techniques,
manipulated content could
rapidly disseminate prejudice and hatred through the Internet,
severely threading the solidarity and harmony of society and
impeding the spread of truth.
For instance,
the news on ``egg shortage caused by avian flu and inflation''
was manipulated to be caused by ``RNA technology''
in order to discredit the research and practices~\cite{newsexp-egg}.
%

\begin{figure}[t]
	\centering
	\includegraphics[width=1\linewidth]{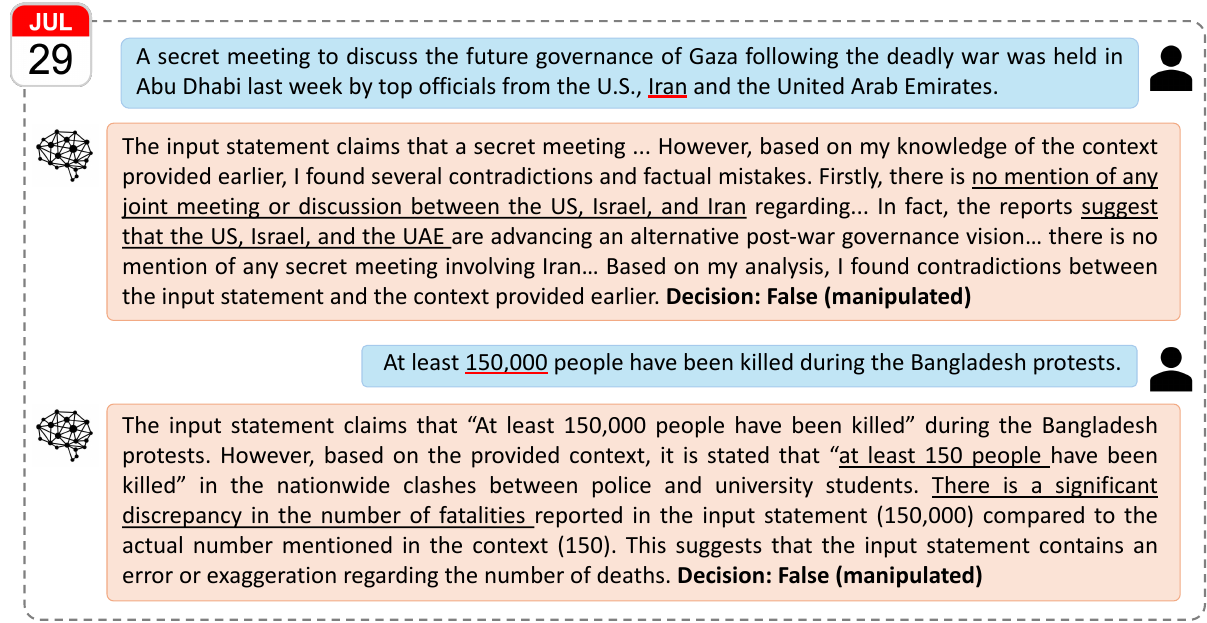}
	\caption{Two examples on
		how \toolname detects and analyzes simulated zero-day manipulated content
		as of 2024 July 29, with
		manipulated content in the user queries underlined in red, and
		key explanations underlined in black.}
	\label{fig:running-example}
\end{figure}

Fortunately,
plenty of efforts have been put into the research on
effective manipulated content detection
in the past decade,
which helps suppress the spreading of disinformation~\cite{politifact,factcheckorg,googlefactcheck,snopes}.
%
%
Early works~\cite{nie2019combining,ma2019sentence,popat2018declare,wu2021evidence}
mainly resort to training machine learning (ML) classifiers
of various architectures to determine
whether a user's claim is truthful.
In addition to simply producing a binary label,
more recent works~\cite{lee2021towards,wang2023explainable,wang2024explainable}
further leverage emerging large language models (LLMs)
to perform a more comprehensive inference and
provide a human-readable justification for the classification,
demonstrating promising performance in
validating historical events and traditional fact-checking tasks.

Nonetheless,
existing solutions mainly
rely on the knowledge learned during the training phase
to support subsequent automatic reasoning and decision-making~\cite{wang2024explainable,lee2021towards,wang2023explainable}.
While some works can take additional context
to determine if a claim is false~\cite{shu2019defend,lu2020gcan},
they require users to manually source, filter, and summarize
such additional information.
The absence of automated real-time knowledge sourcing about
ongoing or recently happened events around the world
makes existing solutions incapable of handling
\emph{zero-day} disinformation caused by manipulated content,
leaving \emph{push-button} manipulated content detection
still an open question in the research community.

One possible reason why real-time contextual information
is not emphasized in prior works is that
these solutions do not explicitly distinguish
manipulated content,
which is still based on true information,
versus entirely false and baseless fabricated content
(which can still be determined to be false
based on logic, common sense, and facts in the knowledge base).
Therefore,
a generic detector that handles
both manipulated content and fabricated content
may not assume the availability of contextual information.
However,
this also means that they will have to forgo the chance of using
relevant contextual information to decide whether a claim is truthful.

In this paper,
we emphasize that contextual information is crucial for a
\emph{push-button} solution to detect and explain
\emph{zero-day} manipulated content,
i.e., fake news created based on events not available
during the training of the detector.
In fact,
real-time information is sometimes even necessary
in deciding the veracity of texts
about ongoing or recently happened events,
as many events simply cannot be inferred or predicted
based on widely accepted ground truth and consensus
(i.e., the black swan theory~\cite{blackswan}).
%
%
%
This is an inherent disadvantage to existing works
that classify a claim merely based on pre-trained outdated data,
as they are incapable of precisely recognizing a piece of
zero-day information,
regardless of how large an ocean of knowledge is used
in the model training phase.
Biden's withdrawal from the 2024 presidential election
is a representative example of an unexpected event
that cannot be verified or reasoned through without access to real-time information~\cite{bidennews}.

However,
even with real-time context information,
detecting manipulated content autonomously is still challenging.
%
%
Manipulated content is often formulated
with minor or sometimes even inconspicuous alterations
based on real stories,
such as the substitution of
persons, time, and venues, or exaggeration of key numbers -- without structural changes in the story.
The minor distortion makes fake news not only easy to mislead people but also difficult to capture by conventional ML models that are widely used in early works~\cite{ma2019sentence,lu2020gcan,nie2019combining,shu2019defend}.
However,
contextual information relevant to this story can be rich.
In a dataset of \allfake manipulated news headlines, each headline contains 126.7 characters on average, with 28.0 characters representing modifiable context, while only 8.4 characters are modified from the original news to mimic malicious manipulation in the real world (see~\autoref{sec:dataset} for dataset details).
%
%
%
This makes
locating the minor distortion
based on a large volume of contextual information
akin to finding a needle in a haystack.



To address these challenges,
we propose \toolname,
a \underline{mani}pulated \underline{co}ntent \underline{d}etector
that can also provide a textual explanation of
why it makes the decision.
More specifically,
\toolname automatically retrieves real-time contextual information from the Internet,
based on which it analyzes the veracity of users' queries accordingly.
In addition to merely producing a true-or-false label or a confidence score,
\toolname also aims to provide explanations for the decision in natural language,
including a pinpoint on the part of the user's input where it suspects
might be manipulated if applicable.

We design \toolname as a two-phase process, namely
\emph{online knowledge retrieval} and
\emph{knowledge-grounded inference by LLMs}.
Specifically,
\toolname first searches the input statement
via mainstream search engines on the Internet
to source the most relevant information.
The search results of the user's query are treated as
contextual knowledge
to support the subsequent reasoning.
Next, we leverage LLMs' capabilities in
natural language comprehension and generation
to infer the veracity of the user's query and
to provide an explanation.
By augmenting the LLM's intrinsic knowledge base
with contextual knowledge,
the LLM would be capable of reasoning about
any piece of information from the query,
even about a recent event.
The veracity inference, at its core,
is to ask the LLM to look for
contradictions, altered text, and/or factual errors
in the claim.
\autoref{fig:running-example}
depicts two running examples of disinformation pieces
and the output of \toolname.

%

Considering mainstream LLMs all enforce constraints
on the maximum token numbers in each inference session,
we resort to retrieval-augmented generation (RAG) techniques
to vectorize the retrieved knowledge and
embed them into the session context of LLMs.
Thus, our approach can take advantage of the powerful capabilities of LLMs
in natural language comprehension and generation
to analyze users' input and explain the decisions made.
At the same time,
the adoption of RAG regulates LLMs to focus on the provided context
during the inference, thereby minimizing the 
effects caused by
hallucination or inconsistent training data~\cite{bechard2024reducinghallucinationstructuredoutputs,shuster2021retrievalaugmentationreduceshallucination}.


While useful,
existing datasets are limited to trivial facts or
are sourced from social-media scenarios~\cite{google2024bigbench,thorne2018fever}
which might not be the best benchmarks to evaluate \toolname
as we target real-world news headlines.
Therefore,
besides comparing with these datasets,
we also propose a dataset
containing manipulated content based on recent news headlines.
Our evaluation shows that
\toolname significantly outperforms existing approaches
in diverse fact-checking and fake news detection tasks.
%
%
The experimental results demonstrate the outstanding performance of \toolname in
detecting zero-day manipulated content about recently happened events around the world,
with an overall F1 score of \ourfone.

\paragraph{Summary}
Our work makes the following contributions:
\begin{itemize}
\item \textbf{An explainable detector for manipulated content}. 
We propose \toolname,
an open-sourced end-to-end autonomous manipulated content detector
capable of sourcing real-time information from the Internet and
recognizing manipulated contents fabricated based on recent real-world events.
Our approach leverages LLMs for inference and explanation generation
without dedicated training or fine-tuning.
To the best of our knowledge,
we are the first of this kind to address the growing concern of zero-day manipulated content specifically.

\item \textbf{A dataset of manipulated news headlines}.
We observe two common types of content manipulation and
follow them to produce a dataset of ``fake news''
by simulating the malicious manipulation based on real-world events.
Our dataset is based on \alltruenews real-world news headlines in English and
contains \allfake pieces of manipulated news headlines,
spanning 20 days in 2024.
The goal of this dataset is also
to encourage more research for a safer Internet environment.

\end{itemize}

Both \toolname and the dataset will be open-sourced
upon the acceptance of this paper\footnote{Available online at \href{https://github.com/cyberooo/manicod}{https://github.com/cyberooo/manicod} for anonymous review.}.
An ethics statement of this work can be found in~\autoref{app:ethics}.

\section{Background}

\subsection{Disinformation and Misinformation}
\label{sec:terms}

In this paper, we take the view that
\emph{disinformation} describes content that is intentionally false and designed to cause harm
while \emph{misinformation} refers to false content created without malicious intents~\cite{ITSAP.00.300,zurko2022disinformation}.
According to the taxonomy in~\cite{wardle2020understanding},
mis- and disinformation can be classified into the following types: 
(1) Fabricated content: new content that is 100\% false. 
(2) \textbf{Manipulated content}: distorted genuine information.
(3) Imposter content: impersonation of genuine sources.
(4) False context: factually accurate content presented together with false and seditious contextual information (e.g., a false text posted with genuine pictures on social networks)
(5) Misleading content: misleading use of genuine information (e.g., partial quotes or cropped pictures)
(6) False connections: When headlines, visuals, or captions do not support the content (e.g., ``clickbait'' content)
(7) Satire and parody: information clearly marked as false but presented in a way as if it were the truth.

As highlighted in the list,
this paper focuses primarily on manipulated content
and our tool \toolname does not aim to detect
other types of mis- and disinformation.
We also do not attempt to further classify whether the manipulated content is misinformation or disinformation,
as such classification would require a tool to infer the motivation behind the production and dissemination of false information,
which is a different research area.

We also assume the
mis- and disinformation circulating on the Internet
is unlikely completely fabricated without
being partially backed by some truth~\cite{Kaliyar2021,Gravanis2019}.
%
Instead, we focus on disinformation in the form of a text that contains minor or indistinguishable perturbations, with all remaining parts being truth to circumvent mainstream fact-checkers~\cite{mazurczyk2024disinformation}.



\subsection{Problem Definition}
\label{sec:problem_definition}

As discussed in~\autoref{sec:terms},
we focus on detecting false information
that is derived from genuine information
by adding minor or even indistinguishable perturbations,
rather than completely ungrounded fabricated content. 
More specifically,
while genuine information can be manipulated in arbitrary ways,
we consider the two most common types of transformations in practice are
\emph{sentiment reversal} and \emph{context alteration}~\cite{Alibai2019}.

Given a truth about a real event as the input,
written as a pair of text and veracity $\langle x, \text{true} \rangle$,
\emph{sentiment reversal} transforms it into a piece of disinformation $x_n$
by negating the sentiment,
while \emph{context alteration} aims to identify the key components
that constitute the context of the input statement
(e.g., appointment and name of persons, date and time, geography, numbers, and quantity)
and replace at least one of them either manually or
through sophisticated AI techniques,
resulting in a piece of disinformation $x_c$.
The two examples illustrated in~\autoref{fig:running-example}
are produced by context alteration,
among which the first one has the country name been tampered with and
the second one has the population of death cases exaggerated by 100 times. 

We regard manipulated content detection as an automatic procedure that
takes a piece of text from the user as the input,
written as $x$, and then outputs a label $y \in \{\text{false}, \text{true}\}$, which corresponds to a ``manipulated'' or ``truthful'' decision.
Suppose there is a dataset containing a set of true statements $X$,
a group of manipulated content created
by reversing the sentiments of true statements written as $X_n$, and
a group of manipulated content created by altering context $X_c$,
an ideal manipulated content detector $f$ should be able to
precisely predict the veracity of these statements.
We define this process as follows:
\vspace{-1em}
\begin{equation}
\begin{split}
\label{equ:detection}
f: x\rightarrow y \,|\, & \left(\left(\forall x \in X, y=\text{true}\right) \right. \\ 
& \left. \, \lor \left(\forall x \in \{X_n, X_c\}, y=\text{false} \right)\right)
\end{split}
\end{equation}
In addition to predicting the veracity,
we note that an auditable manipulated content detector should
ideally be capable of explaining or justifying the predicted veracity
according to the knowledge either learned during the training process or
cited from reliable sources on the Internet.
In this work,
we assume a satisfactory explanation should at least
identify the words or phrases that constitute the manipulation and
consequently result in the false information.
Especially for the cases produced by context alteration,
we expect the detector could produce an explanation
containing the key item that has been altered,
even if that key item does not appear in the user input
(e.g., ``Israel'' in the first running example in~\autoref{fig:running-example}).

\begin{figure*}[t]
	\centering
	\includegraphics[width=0.9\linewidth]{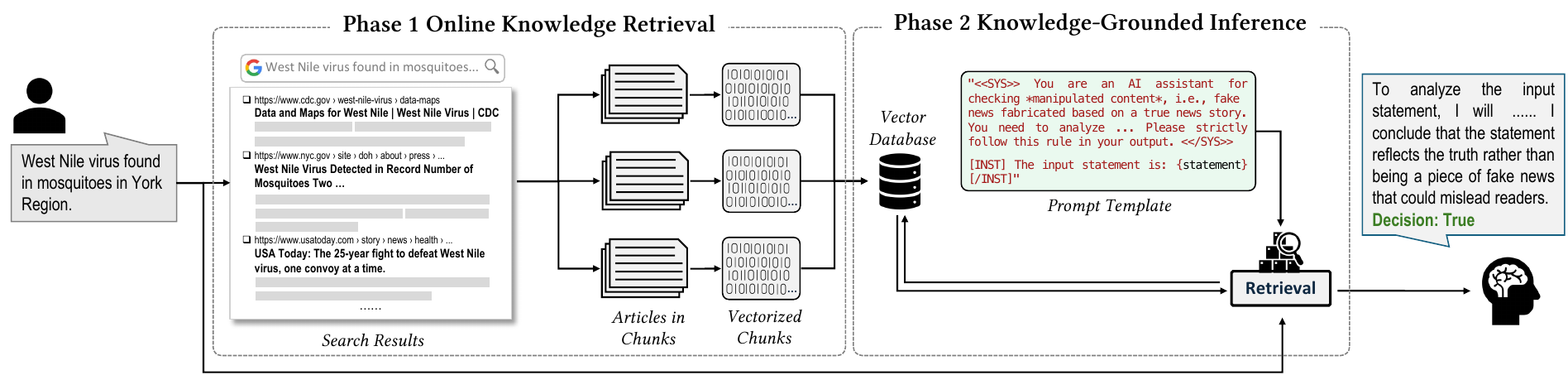}
	\vspace{-0.3em}
	\caption{An overview of our disinformation detection framework \toolname}
	\label{fig:overview}
\end{figure*}

\subsection{Related Work}
\label{sec:related_work}

\paragraph{Textual Veracity Assessment by Conventional ML Models}
The detection of manipulated content
has been studied in the research community for decades.
Early approaches mostly leverage conventional ML classification to predict
if a statement is true or false
either through the lens of \emph{claim verification},
which describes the tasks to determine the veracity of a textual claim through a list of relevant comments or evidence.

%
For example, a piece of false information
is widely regarded to have obvious differences
with truth in terms of
salient phrases~\cite{popat2018declare,wu2021evidence} or
news attributes~\cite{yang2019xfake}. 
In addition to merely making predictions,
the research community has also put persistent effort into producing explanations.
Shu et al.~\cite{shu2019defend}
adopts a hierarchy neural network model to
evaluate the veracity of a claim and
meanwhile nominate the top-$k$ check-worthy sentences
from the accompanying comments for explanation purposes,
thereby achieving explainable fake news detection.
Ma et al.~\cite{ma2019sentence}
leverage hierarchical attention networks to
propose sentence-level embedding for claim verification,
and accordingly highlight the embedding suspicious of falseness
for explanation purposes.
Lu and Li~\cite{lu2020gcan}
adopt a graph neural network that encodes Twitter posts and comments
to determine whether the posts on social networks are fake news.
Nie et al.~\cite{nie2019combining}
investigate the usage of neural semantic matching networks
in fact extraction and veracity verification. 

In summary,
prior works that resort to conventional ML classifiers
have a common limitation as they can only evaluate the veracity of a textual input
along with curated supporting documents,
and their explainability is often restricted to
nominating some of the most relevant documents or sentences that
still require users to manually connect the dots,
which is far from a practical detector for zero-day manipulated content.

\paragraph{Disinformation Detection with LLMs}
After the debut of the BERT model family,
the research community started exploring the adoption of LLMs in
fact-checking and disinformation detection in general.
Lee et al.~\cite{lee2021towards}
train a few BERT-based LLMs for fact-checking,
which are evaluated on two new COVID-related datasets,
demonstrating the powerful capabilities of LLMs
in claim reasoning for the first time.
Wang et al.~\cite{wang2023explainable}
leverage the chain-of-thought mechanism of LLM
for fact verification.
Specifically,
it allows users to provide relevant documents
as the context through the conversation with LLMs,
and then resorts to the backend LLMs to reason the veracity based on the given context.
Wang et al.~\cite{wang2024explainable}
propose a defense-based explainable fake news detection solution,
which uses an evidence-extraction module to split the relevant knowledge,
referred to as ``wisdom of crowds'' in the literature,
into two competing parties and
leverages LLM to reason separately for the veracity of the input claim.
%
By doing so,
it significantly reduces the reliance on the quality of knowledge and
minimizes the impact of occasional inaccurate or biased information from the knowledge.

However,
the effectiveness of existing LLM-based approaches
is limited to scenarios where ground truth data is available,
either from the intrinsic knowledge base of the LLMs or
supplied by users manually.
%
%
%
Furthermore,
considering all mainstream LLMs can only take a limited number of tokens
to build the context,
it is technically hard to supply a large amount of knowledge
through in-context learning~\cite{wang2023explainable}.
Niu et al.~\cite{niu2024veract} address this challenge through LLM based retrieval-augmented detection but rely on fine-tuning open-sourced models or closed-sourced ones like GPT, hindering reproduction without powerful hardware infrastructure or financial budget.
To the best of our knowledge,
our approach is the first of its kind to achieve
automated large-scale context sourcing and embedding based on a pre-trained open-sourced LLM,
making it a fully autonomous detector for zero-day manipulated content.

\paragraph{Fairness Evaluations}
Combating disinformation is a systematic work and
existing solutions are also evaluated by
other research fronts (e.g., usable security).
We discuss these works in~\autoref{app:related-work}.


\section{Our Approach}
\label{sec:overview}

A summary of the proposed \toolname is shown in~\autoref{fig:overview}. 
\toolname adopts a two-phase workflow,
namely the \emph{online knowledge retrieval} phase and
the \emph{LLM-based knowledge-grounded inference} phase.
We detail these two phases in the rest of this section.

\paragraph{Online Knowledge Retrieval}
In the first phase,
our approach takes users’ claims as the input and
searches them via mainstream search engines (e.g., Google)
to source relevant information.
Although \toolname can detect manipulated content generated
based on historical events,
our approach aims to also detect zero-day manipulated content
on recent even ongoing events,
for which analyzing it merely based on the knowledge of LLM or
relying on an additional training process with an off-the-shelf dataset is infeasible.
For that reason,
this step aims to source a significant number of relevant documents autonomously
and construct a context out of the retrieved information.
The context contains necessary knowledge about the input claim
to facilitate later inference and justification. 
While we try to crawl from credible sources only
(e.g., Alexa top 1 million),
we remark that documents
released by even reputable sources,
may still turn out to be false or controversial on a later date.
For example,
The New York Times published 278 corrections for its online news
in September 2024~\cite{nytimecorrections}.
To simplify our study,
we assume the knowledge retrieved from the Internet is true.

Implementation-wise,
we resort to SerpAPI~\cite{web:serpapi} to search the users' claims on Google and
collect the top-$k$ search results from the search list,
i.e., the URLs, based on the relevance ranking performed by Google.
%
We then adopt BeautifulSoup~\cite{web:beautifulsoup}
to crawl the text from the URLs.
The crawled text of each webpage will be saved locally as a document.
In case some websites have applied anti-crawling technologies,
which we did experience with a few notable global news websites, e.g., Reuters,
we skipped that URL and sequentially moved to the next one
to ensure we source the $k$ most relevant webpages to construct the context. 

After the documents are saved,
we prepare for feeding them to the next phase:
LLM-based inference.
Intuitively, we aim to pass all collected documents as a whole,
together with the user's input claim,
to the conversation with LLM to construct the context.
This process is referred to as
in-context learning and has been adopted in~\cite{wang2023explainable}.
However,
\toolname differs from prior work as
it is designed to run autonomously with self-crawled raw textual content
instead of well-organized off-the-shelf processed documents.
%
%
It may contain too much irrelevant information,
such as navigating web components and advertisements,
mixed into the actual contents that \toolname looked for.
This irrelevant content will overwhelm the LLM
to exceed the strict token limitation and
threat the scalability of our approach
if a larger number of relevant web pages is required.
%

To tackle this challenge,
we adopt the RAG technique to encode the relevant information into a vector database. 
Specifically, we resort to an open-sourced toolkit named LangChain~\cite{web:langchain2024langchain},
with nomic-embed-text~\cite{web:nomic-embed} model
to vectorize the plain text of the documents,
and ChromaDB~\cite{web:chroma2024chroma} as the choice of vector database.
Our approach is implemented in Python and
the RAG is realized based on the popular open-sourced and LLM-hosting framework Ollama~\cite{ollama2024ollama}
%
At the end of this phase,
the top-$k$ available documents are vectorized
into a temporary database and
augmented into the LLM's query session of \toolname.  

\paragraph{Knowledge-Grounded Inference}
With the RAG set up,
the second phase of \toolname aims to
determine the veracity of the input claim
by analyzing the augmented context sourced from the Internet and explaining the decision in natural language.
We note that \toolname should work with any LLMs since the veracity inference is mainly based on the external context rather than the intrinsic knowledge. 
Still, we have conducted a small scale experiment to compare the performance of our approach with different mainstream state-of-the-art LLMs. We find that the Meta's Llama 3.1 8b Instruct performs the best among three other models, and as a result, we adopt Llama 3.1 in this phase. We present more details about the model selection in~\autoref{app:model_selection} to save space.

Intuitively,
we shall only need to draft a prompt template containing
an instruction specifying the requirement causally
(i.e., determining whether the given claim is false)
followed by the user's input.
However, during our study,
we find that requiring LLMs to produce a reasonable decision and explanation
turns out to be a non-trivial process. 
During our pilot experiment,
we find that the adopted LLM tends to produce biased decisions
that favors the user.
For example,
the LLM often determines the claim as ``false information''
if we asked ``whether the claim is a piece of false information?'' in the instruction.
Conversely,
the LLM tends to answer ``truth'' if we ask ``whether the claim is a truth?''
%
%
We found this phenomenon becomes more severe
when we are testing with truth claims,
from which the LLM might struggle to identify the trace of any factual mistakes and
eventually chose to favor the user's will~\cite{naveed2024comprehensiveoverviewlargelanguage}.

\begin{figure}[t]
\includegraphics[width=1\linewidth,trim={3pt 1pt 3pt 1pt},clip]{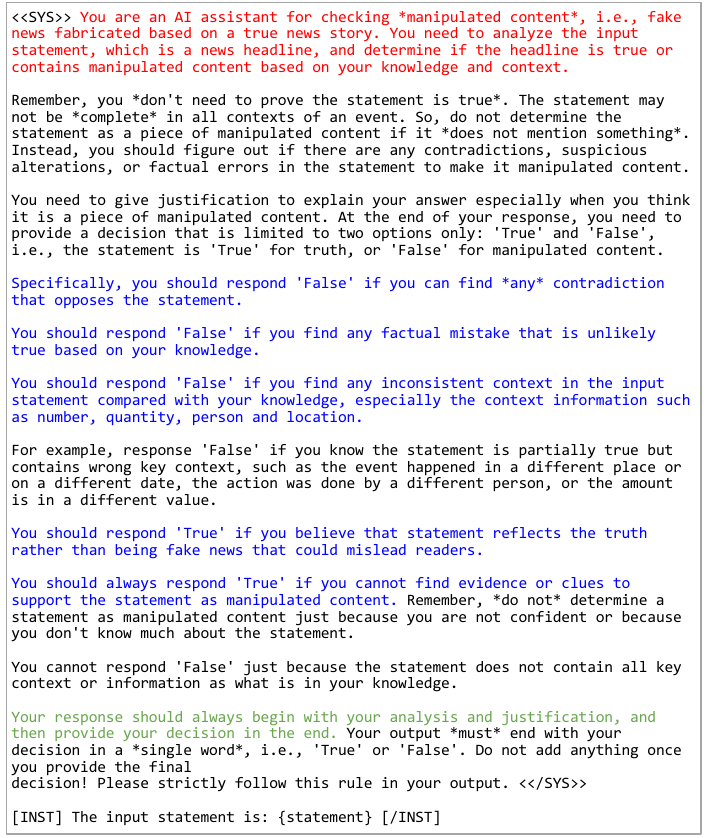}
\vspace{-2em}
\caption{The prompt template used in \toolname
\textnormal{with task description, key inference rules, and output instructions
highlighted in red, blue, and green, respectively}.}
\label{fig:prompt}
\end{figure}

To tackle this issue,
we deliberately designed our prompt template to minimize LLM's
bias in always producing decisions that might favor the user.
We instruct the LLM to reason the input statement first
(e.g., what a piece of manipulated content looks like,
and in which case a true or false decision should be produced)
and make the binary prediction in the end,
to minimize the occurrence that
the adopted LLM makes the prediction
merely based on guessing the user's intention.
The creation of prompt template used for \toolname
is a continuous tuning process
until we find an optimal balance of true and false predictions,
i.e., we repeatedly instruct LLMs with ``hard'' cases
that tend to result in different predictions until
we find a prompt that is stable and yields a reasonable performance.
The final prompt can be found in~\autoref{fig:prompt}.
%

\section{A Dataset of Manipulated Content}
\label{sec:dataset}

As \toolname is a detector specialized for manipulated content,
its effectiveness is best evaluated by a dataset that
simulates how real-world news headlines will be manipulated in reality.
However, to the best of our knowledge,
most existing open-access datasets are either
designed for checking trivial facts~\cite{google2024bigbench}
that can be determined using LLM intrinsic knowledge or
limited in the context of social network posts~\cite{thorne2018fever}.
There does not exist an off-the-shelf dataset of manipulated real-world events
that suits the objective of \toolname to reflect everyday zero-day fake news.
We thus propose a dataset for manipulated content for evaluation purposes and
also to benefit future works of similar purposes.

\begin{figure}[t]
	\centering
	\includegraphics[width=1\linewidth]{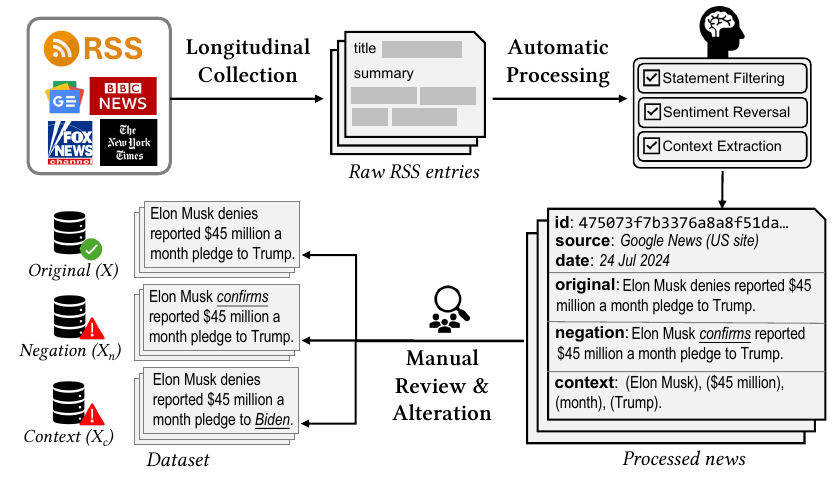}
	\vspace{-2em}
	\caption{An overview of our dataset creation}
	\label{fig:dataset}
\end{figure}

Our dataset is created following the pipeline in~\autoref{fig:dataset}.
The first step is to collect news headlines longitudinally.
Specifically,
we searched notable English news websites with free RSS subscriptions and
identified four sources: Google News, BBC, New York Times, and Fox News,
from which all news entries are crawled.
For each entry in the RSS feed,
we concatenate the news title and summary (if available)
as the news headline. 
We also collect the date of the news being released and
the region or countries of the news site into the dataset
to help customize the online searching
(i.e., Phase 1 in~\autoref{sec:overview}).
Our crawling started on 24 July 2024 and lasted for 20 days.

Next, we resort to the same LLM used in our knowledge-grounded inference
(see Phase 2 in~\autoref{sec:overview}) to process the collected RSS entries
with three steps automatically:
%
(1) We instructed the LLM to filter out news headlines
that are not informative or self-contained to be a claim or
a statement that is applicable for veracity assessment,
such as a question sentence ``\textit{Are we in a summer COVID wave?}''\footnote{A news headline from BBC Health site on 31 Jul 2024.} or an incomplete statement ``\textit{Here are the Daily Lotto numbers}.''\footnote{A news headline from Google News South Africa on 28 Jul 2024.}
(2) We then asked the LLM to produce a negation of the news headline
by identifying and reversing the sentiment in the text. 
The negations of the news headlines will constitute the negation set ($X_n$ in~\autoref{sec:problem_definition}). 
(3) We also leverage LLMs to extract the key contexts from the news headline,
including
persons' names and titles,
geographical terms such as country, state, and city,
quantity, and units.
These extracted contexts will be used as ingredients
to be added back to the original news headlines
to form a separate set of manipulated content by context alteration (i.e., $X_c$).
We present the detailed prompts involved in automatic processing in~\ref{app:appendix-dataset-prompt}.

Finally, we manually review and adjust the contextual information to ensure the authenticity of the dataset.
Specifically, three team members with previous experience in combatting manipulated content or reading news daily manually reviewed all the generated news headlines and replaced the original context with the altered ones to produce simulated fake news containing manipulated content.
For the generated negations, our manual review filtered out inapplicable news or improper negations generated by LLMs.
A typical example inapplicable negation is provided in Appendix~\ref{app:appendix-dataset-negation-example}.

We remark that manual effort is necessary for context alteration
to maintain a high quality dataset
because we need to ensure the fake news with altered context
contradicts the ground truth. 
For example,
given the original news headline
``\textit{At least 150 people have been killed in Bangladesh protests}''\footnote{A news headline from NYT World news on 26 Jul 2024.},
the production of fake news by altering the casualty number
must be a number greater than 150, otherwise,
the altered version should still be considered as truth. 
Our manual effort also ensures the logical correctness of the manipulated content and prevents the altered content from becoming another truth. 
For example, given a news headline
``\textit{Ukraine wins its first medal in Paris Olympic}''\footnote{A news headline from Google News Australia on 30 Jul 2024.},
replacing the word ``Ukraine'' with a few widely discussed countries
like USA or China would not make it manipulated content. 
Instead,
a good candidate to replace the country name in the given context
would be a country without any medal won from the Olympics (e.g., Mexico),
or a country that has its first medal won only after the date of the news
(e.g., Singapore, with its first medal won on a later date on 9 Aug). 
%


\begin{figure}[t]
\tikzstyle{every node}=[font=\Large]
\resizebox{\linewidth}{!}{%
\begin{tikzpicture}
	\pie{ 
		9.2/\makecell[t]{Canada (230)}, 
		8.6/\makecell[t]{Singapore (214)}, 
		8.2/\makecell[t]{Australia (205)}, 
		7.0/\makecell[t]{South Africa (176)}, 
		5.9/\makecell[t]{New Zealand(147)},
		21.3/\makecell[t]{UK (533)}, 
		39.8/\makecell[t]{US \& Worldwide (995)}
		}
\end{tikzpicture}
\hspace{0.0cm}
\begin{tikzpicture}
	\pie{ 
		17.8/\makecell[t]{Google News\\(1444)}, 
		13.1/\makecell[t]{BBC (443)}, 
		11.4/\makecell[t]{New York\\Times (328)}, 
		57.8/\makecell[t]{Fox News (285)}
	}
\end{tikzpicture}
}
\vspace{-0.45em}
\caption{Distribution of the 2,500 news collected by regions (Left) and by news providers (Right) \label{fig:news_dist}}

\end{figure}
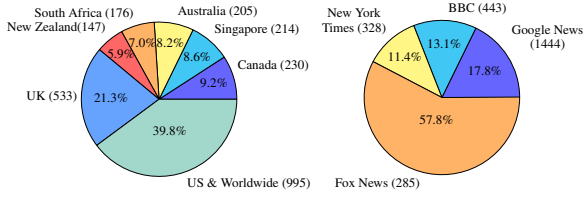

As a result, we collect \allnewscollected valid news headlines. 
The distribution of the news collected is presented in~\autoref{fig:news_dist}.
The original version of these news headlines will constitute the truth set ($X$) in our dataset.
For the manipulated content set, we managed to produce \allfakeneg and \allfakectx fake news containing manipulated content by sentiment reversal ($X_n$) and context alteration ($X_c$), respectively. 
We will present our evaluation of this dataset in~\autoref{sec:results}.
Due to the space limitation, we provide more details of the context alteration and discuss potential bias in Appendix~\ref{app:appendix-dataset-context}.

\section{Evaluation}
\label{sec:evaluation}

In this section, we report the performance of \toolname. 
To explore the effectiveness of the knowledge-grounded inference, we try to answer the following research questions (RQs):
\begin{itemize}
    \item RQ1: How does \toolname perform on our proposed manipulated contents dataset? 
    \item RQ2: How does augmenting real-time knowledge help in identifying manipulated contents? 
    \item RQ3: How robust is \toolname in detecting other misinformation, as presented in~\autoref{sec:terms}? 
\end{itemize}

\paragraph{Experiment Settings}
We set the temperature of the adopted LLM to $0.1$ for all the testing involved in this work to maintain the consistency of LLM outputs and reproducibility.
However, we remark that 
the decision produced by the adopted LLM, i.e., the claim is truth or a piece of disinformation, may still change occasionally. 
For that reason, we repeated the testing of each claim three times and recorded the majority results.

For the scale of online knowledge retrieval, i.e., the selection of $k$, we performed small-scale pilot tests and experimentally found that $k=3$ best balances veracity prediction performance and execution time. 
We implement the RAG with \textit{RecursiveCharacterTextSplitter}~\cite{web:langchain2024recursively}. We follow the default settings of the official tutorial and set the chunk size and overlap to 100 and 20, respectively. The number of retrieved chunks\footnote{We note that the number of retrieved chunks is also referred to \textit{top-$k$} in the LangChain documentation, which indicates the number of chunks to be retrieved from the vectorized database and consequently appended to the context of user input. It is different from the $k$ that we have mentioned earlier, i.e., number of documents to be retrieved from the online search results.} is set to 5. 

During the pilot testing, we also observed that the LLM produced non-conclusive results (e.g., ``I don’t know'', at approximately 5\% occurrences).
%
We adopt a conservative stance and treat all non-conclusive results as wrong,
i.e.,
non-conclusive results over truth are treated as disinformation and
non-conclusive results over disinformation are treated as truth.

Lastly, we stipulate that a successful detection needs to precisely identify the manipulated content to make it explainable.
Therefore, we only accept a correct prediction of disinformation caused by context alteration (i.e., claim from $X_c$) if \toolname simultaneously mentioned the original context and its replacement in its output.



\subsection{RQ1: Veracity Prediction Performance}
\label{sec:results}

After evaluating the proposed manipulated content dataset, our approach achieved promising results, with an overall \ourprecision precision and \ourrecall recall. The F1 score of our evaluation is \ourfone. 
We present the confusion matrix of the evaluation in~\autoref{app:confusion_matrix}.

Being specific to each type of claim, our approach performs the best on verifying manipulated content by reversed sentiment, i.e., $X_n$, evidenced by the accuracy of \negationacc, followed by manipulation by context alteration ($X_c$) and original headlines ($X$), with accuracy at \contextacc and \indexacc, respectively. 
Although \toolname achieves promising overall results, we find the veracity prediction of the original news headlines worse than the remaining two types of claims. 
We suspect such unexpected results may be because the news' or the LLM's intrinsic veracity is not completely true. 
The online searching performed at a later moment than the news being drafted may further amplify the gap between the news and ground truth.
For example, a piece of news that contains ``\emph{Prime Minister Keir Starmer}''\footnote{A news headline collected from Fox News on 5 Aug 2024.} is considered as False as ``Rishi Sunak has been serving as the Prime Minister of the United Kingdom since October 2022, and Keir Starmer is actually the leader of the Labour Party in the UK'', even the search results also highlights that the current UK Prime Minister is Keir Starmer.

Finally, we also record the time lapsed during the evaluation. We present more details in~\autoref{app:appendix-time-lapsed} to save space.

\subsection{RQ2: Ablation Study}
\label{sec:ablation-study}

In addition to evaluating \toolname on our manipulated content dataset, we also conducted an ablation study on the effect of the knowledge-grounded inference based on online retrieval.
Specifically, we compare our approach with the two popular LLMs: the LLM adopted in \toolname named Llama3.1 (8b), and GPT-4o-mini, known as the most powerful proprietary LLM at the moment of this paper being drafted.
Both LLMs are tested with the same dataset but without augmenting the context retrieved from the Internet. Given the latest data trained for the two LLMs are far earlier than the news included in our dataset at this moment,\footnote{The knowledge cutoff dates of the Llama 3.1 (8b) and GPT-4o-mini models are Dec. 2023 and Oct. 2023, respectively.} we can assume both LLMs can only predict the claims jointly based on the outdated knowledge and the language logic in the text. 


\begin{table}[t]
\centering
\def\arraystretch{1.01}
\setlength{\tabcolsep}{3.4pt}
\caption{Comparison of our approach with LLMs without online knowledge retrieval\label{tab:ablation_study}}
\vspace{-1em}
\small

\resizebox{\linewidth}{!}{%
\begin{tabular}{lrrr}
\hline
 & \multicolumn{3}{c}{Accuracy} \\ \cline{2-4}
& Original ($X$) & Negation ($X_n$) & Context altered ($X_c$) \\ \hline 
Llama 3.1 (direct) & \llamaindexacc & \llamanegationacc & \llamacontextacc   \\
GPT-4o (direct)    & \gptindexacc & \gptnegationacc & \gptcontextacc  \\ \hline 
\textit{Ours} (w. Llama 3.1)  & \textbf{\indexacc} & \textbf{\negationacc} & \textbf{\contextacc}  \\ \hline
\end{tabular}
}

\end{table}

The results of our ablation study are presented in~\autoref{tab:ablation_study}. We can observe both LLMs without the context constructed from online knowledge perform poorly on all three categories of claims. This implies the effectiveness of our two-phase design for real-time manipulated content detection.

\subsection{RQ3: Comparison to Other Methods}
\label{sec:comparison-baseline}

Our proposed dataset contains only slightly manipulated content based on real-world events. 
That makes the performance of \toolname in handling other types of disinformation unanswered. 
For that reason, we benchmarked our approach with existing datasets for claim verification and fact-checking and compared the performance of \toolname with previous research.


\begin{table}[t]
\centering
\def\arraystretch{1.01}
\setlength{\tabcolsep}{1pt}
\caption{Veracity prediction results on binary-class fact-checking tasks, with our approach compared with the state-of-the-art approaches \label{tab:benchmarking_binary_sota}}
\vspace{-0.6em}
\footnotesize
\resizebox{\linewidth}{!}{%
\begin{tabular}{lccHcc}
\hline
 & \multicolumn{2}{c}{COVID~\cite{google2024bigbench}} & \hspace{0.2cm} & \multicolumn{2}{c}{FEVER~\cite{thorne2018fever}} \\ \cline{2-3}\cline{5-6}
 & Accuracy      & F1 score   &  & Accuracy  & F1 score    \\ \hline 
\multicolumn{6}{l}{\textit{Traditional approaches}} \\ \hline
\hspace{0.03cm} XLNet\textsubscript{\textit{ft}}~\cite{lee2021towards} & 0.632    & 0.520 & & 0.492 & 0.484 \\ \hline 
\multicolumn{6}{l}{\textit{Perplexity-based LLMs}} \\ \hline
\hspace{0.03cm} BERT\textsubscript{\textit{PPL}}~\cite{lee2021towards} & 0.625    & 0.611 & & 0.574 & 0.569 \\
\hspace{0.03cm} GPT2\textsubscript{\textit{PPL} (15b)}~\cite{lee2021towards} & {0.783}    & {0.776} & & {0.736} & {0.717} \\ \hline 
\textit{Ours}                            & {{0.873}} & {{0.911}} & & {{0.807}} & {{0.788}} \\ \hline

\end{tabular}
}
\end{table}

Our approach is firstly benchmarked on two datasets designed for binary label fact-checking tasks, namely COVID-Scientific~\cite{google2024bigbench} and FEVER~\cite{thorne2018fever}.
The former collects COVID-19-related myths and truths labeled by reliable sources including WHO and CDC, and the latter is designed for a fact extraction and verification task containing a large number of erroneous information by altering sentences on Wikipedia.
We consider claims from both datasets to be sufficiently verified by online information, and therefore, we do not consider the evidence provided in the datasets. 
We retrieve the existing literature and compare our approach with the best performing existing approach~\cite{lee2021towards}.
The comparison with previous work that benchmarked these two datasets are presented in~\autoref{tab:benchmarking_binary_sota}.


\begin{table}[t]
\centering
\def\arraystretch{1.01}
\setlength{\tabcolsep}{2.5pt}
\caption{Veracity prediction results on multi-class claim verification tasks, with our approach compared with the state-of-the-art in different types and the best results of existing work are shown in \textbf{bold} font  \label{tab:benchmarking_multilabel_sota}}
\vspace{-0.5em}
\footnotesize

\resizebox{\linewidth}{!}{%
\begin{tabular}{lcccHccc}
\hline
 & \multicolumn{3}{c}{LIAR-RAW~\cite{wang2017liar}} && \multicolumn{3}{c}{RAWFC~\cite{yang2022coarse}} \\ \cline{2-4} \cline{6-8}
 & Precision      & Recall & F1 score && Precision & Recall & F1 score \\ \hline 
\multicolumn{8}{l}{\textit{Traditional approaches}} \\ \hline
\hspace{0.03cm} dEFEND~\cite{shu2019defend}   & 
0.231          & 0.186  & 0.175    && 0.449     & 0.433  & 0.441    \\
\hspace{0.03cm} SBERT-FC~\cite{kotonya2020explainable} & 
0.241          & 0.221  & 0.222    && 0.511     & 0.459  & 0.455    \\
\hspace{0.03cm} GenFE-MT~\cite{atanasova2020generating}  & 
0.186          & 0.199  & 0.152    && 0.456     & 0.453  & 0.451    \\
\hspace{0.03cm} CofCED~\cite{yang2022coarse} & 
0.295          & 0.296  & 0.289    && 0.530     & 0.510  & 0.511    \\ \hline 
\multicolumn{8}{l}{\textit{LLM-based approaches}} \\ \hline
\hspace{0.03cm} LLaMA2~\textsubscript{claim}~\cite{ouyang2022training}   & 
0.171          & 0.174  & 0.151    && 0.373     & 0.380  & 0.368    \\
\hspace{0.03cm} ChatGPT\textsubscript{claim}~\cite{ouyang2022training}   & 
0.254          & 0.273  & 0.251    && 0.477     & 0.486  & 0.444    \\
\hspace{0.03cm} FactLLaMA\textsubscript{know}~\cite{cheung2023factllama}    & 
\textbf{0.325 }         & 0.321  & 0.304    && 0.561     & 0.555  & 0.557    \\ \hline 
\multicolumn{8}{l}{\textit{Defense-based approaches with LLM Reasoning}}  \\ \hline
\hspace{0.03cm} L-Defense\textsubscript{LLaMA2}~\cite{wang2023explainable}      & 
0.316          & 0.317  & \textbf{0.314}    && 0.610     & 0.600  & 0.601    \\
\hspace{0.03cm} L-Defense\textsubscript{ChatGPT}~\cite{wang2023explainable}     & 
0.306          & \textbf{0.322}  & 0.305    && \textbf{0.617}     & \textbf{0.610}  & \textbf{0.612}    \\ \hline 
\textit{Ours} & 0.920          & 0.910  & 0.915    && 0.821     & 0.932  & 0.873   \\ \hline
\end{tabular}
}

\end{table}

In addition to binary labeled tasks, two more datasets designed for claim verification are considered in our comparison, namely RAWFC~\cite{yang2022coarse} and LIAR-RAW~\cite{wang2017liar}. 
The RAWFC dataset is a collection of claims sourced from Snopes~\cite{snopes} labeled either true, false, or half-true. The LIAR dataset contains over 12,000 fine-grained claims collected from PolitiFact~\cite{politifact}, labeled by six options based on the degree of falseness, i.e., true, mostly true, half-true, barely true, false, or pants-on-fire. 
As \toolname is designed to only produce binary decisions, our comparison does not cover claims with ambiguous or controversial veracity. Specifically, our comparison excludes claims with the two neutral labels, i.e., half-true and barely true, from the two datasets.
Unlike the comparison with COVID and FEVER datasets, we mock the online retrieval process by directly augmenting the evidence provided by RAWFC and LIAR-RAW datasets to realize the knowledge-grounded inference.
Still, we review the relevant literature~\cite{shu2019defend,kotonya2020explainable,atanasova2020generating,yang2022coarse,ouyang2022training,cheung2023factllama,wang2023explainable} and compare \toolname with the representative and state-of-the-art existing approaches in different types.
We show our results in~\autoref{tab:benchmarking_multilabel_sota}.

As the results show, our approach outperforms existing approaches on the four datasets all the time.
Compared with the best performing existing approaches on each task, \toolname can bring an improvement of the F1 score up to 17.3\% for binary-label fact-checking tasks (i.e., 0.911 versus 0.776), and an up to 1.9x improvement of the F1 score for multi-label claim verification tasks (0.915 versus 0.314). 

We remark that our approach is  primarily designed to detect manipulated contents based on widely accepted ground truth
and consensus on the Internet despite the benchmarking showing impressive results on datasets of fabricated fake news or rumors such as Snopes\footnote{The official FAQ of Snopes~\cite{snopes2025faq} writes that the verification is jointly realized by ``attempting to contact the source of the claim for elaboration and supporting information'' and ``attempting to contact individuals and organizations who would be knowledgeable about it.'' Its data may contain fabricated contents that can only be clarified by those involved.}. This, again, demonstrates the effectiveness of knowledge-grounded inference and reveals a promising direction in relevant research. 
We also note that our benchmarking does not claim an exhaustive comparison of existing approaches and datasets but focusing on representative ones that have been widely referred to in prior research. 

\section{Conclusions}
\label{sec:conclusions}

We propose \toolname,
an LLM-based manipulated content detector.
Our approach can automatically search and retrieve knowledge from the Internet and,
therefore, can analyze the veracity of zero-day manipulated content
about recent real-world events.
\toolname also leverages the latest open-sourced LLM
to perform a knowledge-grounded inference and
to provide explanations in natural language to justify its decisions.
We also create a manipulated content dataset dedicated to evaluating \toolname
as well as future research efforts on combating manipulated content.
Our evaluation shows that
\toolname can effectively detect zero-day manipulated content
evidenced by an overall F1 score of \ourfone.
A comparison with prior work also demonstrates
\toolname's superior performance in diverse fact-checking and claim-verification tasks. 
Our work advances online safety research and
we hope \toolname can inspire future efforts
on leveraging powerful AI technologies
in pursuit of social goodness.

\section{Limitations}
\label{sec:discussion}

We find a few limitations that may threaten the effectiveness of our approach. We categorize the identified limitations into internal and external ones and discuss them in this section.

\paragraph{Internal Limitations}
First, the online knowledge retrieval of \toolname directly searches the user's claim through a search engine.
Its performance can be downgraded when dealing with lengthy claims that exceed the word count limit or contain complicated contexts that the search engine may not find the best-suited information straightforwardly. 
To tackle this challenge, distilling critical contexts from the original claim as the search keywords may be a promising direction for future improvement. 
Another potential mitigation is decomposing the original long claim into several atomic statements, followed by separate online searching and knowledge augmentation. This has been shown effective in veracity detection in~\cite{niu2024veract}. We aim to explore this in future work.

Second, while the manipulated news in the proposed dataset is carefully crafted to be factually false, it may not fully reflect the complexity of real-world disinformation.
In practice, fakesters carefully blend partial truths with misleading contexts, making detecting them more challenging. 
These real-world manipulations may also follow different distribution patterns, which our datasets may not adequately represent. 
Moreover, fake news in real-world scenarios may be especially designed to evade detection by emerging AI-based detectors and may have the potential to deceive the LLM into making incorrect decisions.

Third, we find that the performance of our approach heavily relies on the choice of LLMs and a high-quality prompt template that fits the adopted LLM. The tuning of an effective prompt template is found to be a non-trivial and time-consuming task. 
For example, we experimentally found that the adopted Llama 3.1 model is more sensitive to terms like ``rumors'' and ``fake news'' rather than ``manipulated content'' and ``disinformation.'' 
For this reason, we have to explicitly ask the LLM to determine if the input claim is ``fake news'' although we are not meant to restrict our scope to this. 
Moreover, we find a fine-grained prompt designed for the adopted LLM does not imply its effectiveness for other LLMs, which may threaten the extensibility of our approach. 
How to effectively instruct LLM to perform complex tasks canonically and generically still awaits our further exploration.

Last, we observe that many false predictions are caused by the mistakes made by LLMs during inference. For example, although we adopted one of the most powerful LLMs, it may still face limitations in certain tasks, including text comprehension, value comparison, and chronological order reasoning. 
Similarly, it was trained with potential bias, incorrect, or outdated data that may be intrinsically prone to generating incorrect results. 
We shall resort to the future improvement of LLMs with more powerful reasoning capabilities to mitigate this concern.
We provide a case study for more details of this discussion in~\autoref{sec:appendix-case-study}.

\paragraph{External Limitations}
The veracity of news headlines from our proposed dataset inspires us to discuss the potential external limitation that may threaten \toolname's validity. 
First, we learn that our collected news, even sourced from reputable websites, may still turn out to be mis- or disinformation, resulting in potential false negative cases (i.e., the original news is assumed to be true but is proven to be false according to online knowledge) in our evaluation. 
Besides that, news about the same event but mentioned on different web pages may conflict with each other, leading to failed reasoning.
A representative example could be the identity of the shooter in the accident of Donald Trump in 2024, for which a piece of fake news saying the shooter is Asian has been widely spreading on the Internet~\cite{xu2024nypost}. 
While our research does not focus on how news should be produced, we can maintain a pool of reputable media to allow online searching only from their domains to maximize the reliability of the online knowledge retrieved.
Although no news website is entirely free of inaccuracies.

Another limitation that affects \toolname's performance is the veracity dynamics of news. During the evaluation, we noticed that knowledge on the Internet may keep changing over time (e.g., the number of causalities may increase, and celebrities' claims may suddenly change).
Although our online search simulates searching on the day the news is released, the web pages with the same URLs may keep updating, so the ground truth may not always be stable. 
Biden's decision to quit the 2024 election would be a notable example to highlight this limitation, as he kept claiming he would continue running for the next president until his announcement on 22 July. 
To mitigate the impact of such limitation, we shall keep updating the datasets with the latest news and evaluate our approach only with the latest data to avoid the unprecedented change of veracity.

\section{Ethics Considerations}
\label{app:ethics}
Our paper focuses on the detection of manipulated real-world fake news with the aim of enhancing the robustness of current large language models (LLMs) in identifying zero-day disinformation—fake news that has not been previously encountered. We aim to develop an accessible solution that enables ordinary individuals to easily verify the authenticity of the content they encounter.

However, we recognize the challenge that fake content creators may adapt and devise countermeasures. Still, compromising our system is non-trivial and would require them to infiltrate major news sources.

\bibliography{reference}

\begin{thebibliography}{48}
\providecommand{\natexlab}[1]{#1}

\bibitem[{Alibašić and Rose(2019)}]{Alibai2019}
Haris Alibašić and Jonathan Rose. 2019.
\newblock \href {https://doi.org/10.1080/10999922.2019.1622359} {Fake news in
  context: Truth and untruths}.
\newblock \emph{Public Integrity}, 21(5):463–468.

\bibitem[{{April Xu}(2024)}]{xu2024nypost}
{April Xu}. 2024.
\newblock New york post falsely claims chinese man shot trump chinese
  communities outraged.
\newblock
  \url{https://www.msn.com/en-us/news/politics/new-york-post-falsely-claims-chinese-man-shot-trump-chinese-communities-outraged/ar-BB1q5Ea9}.

\bibitem[{Atanasova et~al.(2020)Atanasova, Simonsen, Lioma, and
  Augenstein}]{atanasova2020generating}
Pepa Atanasova, Jakob~Grue Simonsen, Christina Lioma, and Isabelle Augenstein.
  2020.
\newblock \href {https://doi.org/10.18653/v1/2020.acl-main.656} {Generating
  fact checking explanations}.
\newblock In \emph{Proceedings of the 58th Annual Meeting of the Association
  for Computational Linguistics}, pages 7352--7364. Association for
  Computational Linguistics.

\bibitem[{Béchard and
  Ayala(2024)}]{bechard2024reducinghallucinationstructuredoutputs}
Patrice Béchard and Orlando~Marquez Ayala. 2024.
\newblock \href {https://arxiv.org/abs/2404.08189} {Reducing hallucination in
  structured outputs via retrieval-augmented generation}.
\newblock \emph{Preprint}, arXiv:2404.08189.

\bibitem[{{Canadian Centre for Cyber Security}(2024)}]{ITSAP.00.300}
{Canadian Centre for Cyber Security}. 2024.
\newblock {How to Identify Misinformation, Disinformation, and Malinformation
  (ITSAP.00.300)}.
\newblock
  \url{https://www.cyber.gc.ca/sites/default/files/misinformation-mesinformation-itsap.00.300-en.pdf}.
\newblock Online; accessed 10 October 2024.

\bibitem[{Chen and Shu(2024)}]{chen2024llmgenerated}
Canyu Chen and Kai Shu. 2024.
\newblock \href {https://openreview.net/forum?id=ccxD4mtkTU} {Can llm-generated
  misinformation be detected?}
\newblock In \emph{The Twelfth International Conference on Learning
  Representations}.

\bibitem[{Cheung and Lam(2023)}]{cheung2023factllama}
Tsun-Hin Cheung and Kin-Man Lam. 2023.
\newblock \href {https://doi.org/10.1109/APSIPAASC58517.2023.10317251}
  {Factllama: Optimizing instruction-following language models with external
  knowledge for automated fact-checking}.
\newblock In \emph{2023 Asia Pacific Signal and Information Processing
  Association Annual Summit and Conference (APSIPA ASC)}, pages 846--853.

\bibitem[{{Chroma}(2024)}]{web:chroma2024chroma}
{Chroma}. 2024.
\newblock Chroma - the open-source embedding database.
\newblock \url{https://github.com/chroma-core/chroma}.
\newblock Online; accessed 04 September 2024.

\bibitem[{{Google}(2021)}]{google2024bigbench}
{Google}. 2021.
\newblock Fact-checking - covid19-scientific.
\newblock
  \url{https://github.com/google/BIG-bench/tree/main/bigbench/benchmark_tasks/fact_checker/covid19_scientific}.
\newblock Online; accessed 04 September 2024.

\bibitem[{{Google}(2024)}]{googlefactcheck}
{Google}. 2024.
\newblock {Fact Check Tools - Google Search}.
\newblock \url{https://toolbox.google.com/factcheck/explorer}.
\newblock Online; accessed 07 October 2024.

\bibitem[{Gravanis et~al.(2019)Gravanis, Vakali, Diamantaras, and
  Karadais}]{Gravanis2019}
Georgios Gravanis, Athena Vakali, Konstantinos Diamantaras, and Panagiotis
  Karadais. 2019.
\newblock \href {https://doi.org/10.1016/j.eswa.2019.03.036} {Behind the cues:
  A benchmarking study for fake news detection}.
\newblock \emph{Expert Systems with Applications}, 128:201–213.

\bibitem[{Kaliyar et~al.(2021)Kaliyar, Goswami, and Narang}]{Kaliyar2021}
Rohit~Kumar Kaliyar, Anurag Goswami, and Pratik Narang. 2021.
\newblock \href {https://doi.org/10.1007/s00521-020-05611-1} {Echofaked:
  improving fake news detection in social media with an efficient deep neural
  network}.
\newblock \emph{Neural Computing and Applications}, 33(14):8597–8613.

\bibitem[{Kotonya and Toni(2020)}]{kotonya2020explainable}
Neema Kotonya and Francesca Toni. 2020.
\newblock \href {https://doi.org/10.18653/v1/2020.emnlp-main.623} {Explainable
  automated fact-checking for public health claims}.
\newblock In \emph{Proceedings of the 2020 Conference on Empirical Methods in
  Natural Language Processing (EMNLP)}, pages 7740--7754. Association for
  Computational Linguistics.

\bibitem[{LangChain(2024)}]{web:langchain2024recursively}
LangChain. 2024.
\newblock {Recursively split by character}.
\newblock
  \url{https://python.langchain.com/v0.1/docs/modules/data_connection/document_transformers/recursive_text_splitter}.
\newblock Online; accessed 04 September 2024.

\bibitem[{{LangChain-AI}(2024)}]{web:langchain2024langchain}
{LangChain-AI}. 2024.
\newblock Langchain.
\newblock \url{https://github.com/langchain-ai/langchain}.
\newblock Online; accessed 04 September 2024.

\bibitem[{Lee et~al.(2021)Lee, Bang, Madotto, Khabsa, and
  Fung}]{lee2021towards}
Nayeon Lee, Yejin Bang, Andrea Madotto, Madian Khabsa, and Pascale Fung. 2021.
\newblock Towards few-shot fact-checking via perplexity.
\newblock \emph{arXiv preprint arXiv:2103.09535}.

\bibitem[{Lu and Te~Li(2020)}]{lu2020gcan}
Yi~Ju Lu and Cheng Te~Li. 2020.
\newblock Gcan: Graph-aware co-attention networks for explainable fake news
  detection on social media.
\newblock In \emph{58th Annual Meeting of the Association for Computational
  Linguistics, ACL 2020}, pages 505--514. Association for Computational
  Linguistics (ACL).

\bibitem[{Ma et~al.(2019)Ma, Gao, Joty, and Wong}]{ma2019sentence}
Jing Ma, Wei Gao, Shafiq Joty, and Kam-Fai Wong. 2019.
\newblock Sentence-level evidence embedding for claim verification with
  hierarchical attention networks.
\newblock Association for Computational Linguistics.

\bibitem[{Mazurczyk et~al.(2024)Mazurczyk, Lee, and
  Vlachos}]{mazurczyk2024disinformation}
Wojciech Mazurczyk, Dongwon Lee, and Andreas Vlachos. 2024.
\newblock \href {https://doi.org/10.1145/3624721} {Disinformation 2.0 in the
  age of ai: A cybersecurity perspective}.
\newblock \emph{Commun. ACM}, 67(3):36–39.

\bibitem[{Naveed et~al.(2024)Naveed, Khan, Qiu, Saqib, Anwar, Usman, Akhtar,
  Barnes, and Mian}]{naveed2024comprehensiveoverviewlargelanguage}
Humza Naveed, Asad~Ullah Khan, Shi Qiu, Muhammad Saqib, Saeed Anwar, Muhammad
  Usman, Naveed Akhtar, Nick Barnes, and Ajmal Mian. 2024.
\newblock \href {https://arxiv.org/abs/2307.06435} {A comprehensive overview of
  large language models}.
\newblock \emph{Preprint}, arXiv:2307.06435.

\bibitem[{Nie et~al.(2019)Nie, Chen, and Bansal}]{nie2019combining}
Yixin Nie, Haonan Chen, and Mohit Bansal. 2019.
\newblock Combining fact extraction and verification with neural semantic
  matching networks.
\newblock In \emph{Proceedings of the AAAI conference on artificial
  intelligence}, volume~33, pages 6859--6866.

\bibitem[{Niu et~al.(2024)Niu, Guan, Wu, Zhu, Song, Zhong, Zhu, Xu, Diao, and
  Zhang}]{niu2024veract}
Cheng Niu, Yang Guan, Yuanhao Wu, Juno Zhu, Juntong Song, Randy Zhong, Kaihua
  Zhu, Siliang Xu, Shizhe Diao, and Tong Zhang. 2024.
\newblock \href {https://doi.org/10.18653/v1/2024.acl-demos.25} {{V}era{CT}
  scan: Retrieval-augmented fake news detection with justifiable reasoning}.
\newblock In \emph{Proceedings of the 62nd Annual Meeting of the Association
  for Computational Linguistics (Volume 3: System Demonstrations)}, pages
  266--277, Bangkok, Thailand. Association for Computational Linguistics.

\bibitem[{{Nomic}(2024)}]{web:nomic-embed}
{Nomic}. 2024.
\newblock {Introducing Nomic Embed: A Truly Open Embedding Model}.
\newblock \url{https://www.nomic.ai/blog/posts/nomic-embed-text-v1}.
\newblock Online; accessed 04 September 2024.

\bibitem[{{Ollama}(2024)}]{ollama2024ollama}
{Ollama}. 2024.
\newblock Ollama.
\newblock \url{https://github.com/ollama/ollama}.
\newblock Online; accessed 04 September 2024.

\bibitem[{Ouyang et~al.(2022)Ouyang, Wu, Jiang, Almeida, Wainwright, Mishkin,
  Zhang, Agarwal, Slama, Ray, Schulman, Hilton, Kelton, Miller, Simens, Askell,
  Welinder, Christiano, Leike, and Lowe}]{ouyang2022training}
Long Ouyang, Jeffrey Wu, Xu~Jiang, Diogo Almeida, Carroll Wainwright, Pamela
  Mishkin, Chong Zhang, Sandhini Agarwal, Katarina Slama, Alex Ray, John
  Schulman, Jacob Hilton, Fraser Kelton, Luke Miller, Maddie Simens, Amanda
  Askell, Peter Welinder, Paul~F Christiano, Jan Leike, and Ryan Lowe. 2022.
\newblock \href
  {https://proceedings.neurips.cc/paper_files/paper/2022/file/b1efde53be364a73914f58805a001731-Paper-Conference.pdf}
  {Training language models to follow instructions with human feedback}.
\newblock In \emph{Advances in Neural Information Processing Systems},
  volume~35, pages 27730--27744. Curran Associates, Inc.

\bibitem[{Popat et~al.(2018)Popat, Mukherjee, Yates, and
  Weikum}]{popat2018declare}
Kashyap Popat, Subhabrata Mukherjee, Andrew Yates, and Gerhard Weikum. 2018.
\newblock Declare: Debunking fake news and false claims using evidence-aware
  deep learning.
\newblock \emph{arXiv preprint arXiv:1809.06416}.

\bibitem[{Post(2024)}]{bidennews}
The~Washington Post. 2024.
\newblock \href
  {https://www.washingtonpost.com/politics/2024/07/21/joe-biden-drops-out/}
  {Biden makes stunning decision to pull out of 2024 race}.

\bibitem[{{Reuters Fact Check}(2023)}]{newsexp-egg}
{Reuters Fact Check}. 2023.
\newblock No evidence ‘rna technology’ in chicken feed behind infertility
  or u.s. egg shortage.
\newblock
  \url{https://www.reuters.com/article/fact-check/no-evidence-rna-technology-in-chicken-feed-behind-infertility-or-us-egg-shor-idUSL1N34N1RJ/}.

\bibitem[{Richardson(2024)}]{web:beautifulsoup}
Leonard Richardson. 2024.
\newblock {Beautiful Soup Documentation}.
\newblock \url{https://beautiful-soup-4.readthedocs.io/}.
\newblock Online; accessed 04 September 2024.

\bibitem[{{SerpApi}(2024)}]{web:serpapi}
{SerpApi}. 2024.
\newblock {SerpApi: Google Search API}.
\newblock \url{https://serpapi.com/}.
\newblock Online; accessed 04 September 2024.

\bibitem[{Shu et~al.(2019)Shu, Cui, Wang, Lee, and Liu}]{shu2019defend}
Kai Shu, Limeng Cui, Suhang Wang, Dongwon Lee, and Huan Liu. 2019.
\newblock defend: Explainable fake news detection.
\newblock In \emph{Proceedings of the 25th ACM SIGKDD international conference
  on knowledge discovery \& data mining}, pages 395--405.

\bibitem[{Shuster et~al.(2021)Shuster, Poff, Chen, Kiela, and
  Weston}]{shuster2021retrievalaugmentationreduceshallucination}
Kurt Shuster, Spencer Poff, Moya Chen, Douwe Kiela, and Jason Weston. 2021.
\newblock \href {https://arxiv.org/abs/2104.07567} {Retrieval augmentation
  reduces hallucination in conversation}.
\newblock \emph{Preprint}, arXiv:2104.07567.

\bibitem[{{Snopes}(2025)}]{snopes2025faq}
{Snopes}. 2025.
\newblock Snopes.com | frequently asked questions.
\newblock \url{https://www.snopes.com/faqs/}.
\newblock Online; accessed 07 October 2024.

\bibitem[{{Snopes Media Group}(2024)}]{snopes}
{Snopes Media Group}. 2024.
\newblock Snopes.com | the definitive face-checking site.
\newblock \url{https://www.snopes.com}.
\newblock Online; accessed 07 October 2024.

\bibitem[{Taleb(2010)}]{blackswan}
N.N. Taleb. 2010.
\newblock \href {https://books.google.com/books?id=7wMuF4A4XF8C} {\emph{The
  Black Swan: Second Edition: The Impact of the Highly Improbable Fragility"}}.
\newblock Incerto. Random House Publishing Group.

\bibitem[{{The Annenberg Public Policy Center}(2024)}]{factcheckorg}
{The Annenberg Public Policy Center}. 2024.
\newblock {FackCheck.org - A Project of The Annenberg Public Policy Center of
  the University of Pennsylvania}.
\newblock \url{https://www.factcheck.org}.
\newblock Online; accessed 07 October 2024.

\bibitem[{{The Poynter Institute}(2024)}]{politifact}
{The Poynter Institute}. 2024.
\newblock {Fact-checks | PolitiFact}.
\newblock \url{https://www.politifact.com/factchecks/list/}.
\newblock Online; accessed 07 October 2024.

\bibitem[{Thorne et~al.(2018)Thorne, Vlachos, Christodoulopoulos, and
  Mittal}]{thorne2018fever}
James Thorne, Andreas Vlachos, Christos Christodoulopoulos, and Arpit Mittal.
  2018.
\newblock \href {https://doi.org/10.18653/v1/N18-1074} {{FEVER}: a large-scale
  dataset for fact extraction and {VER}ification}.
\newblock In \emph{Proceedings of the 2018 Conference of the North {A}merican
  Chapter of the Association for Computational Linguistics: Human Language
  Technologies, Volume 1 (Long Papers)}, pages 809--819, New Orleans,
  Louisiana. Association for Computational Linguistics.

\bibitem[{Times(2024)}]{nytimecorrections}
The New~York Times. 2024.
\newblock \href {https://www.nytimes.com/international/section/corrections}
  {Corrections}.

\bibitem[{Wang et~al.(2024)Wang, Ma, Lin, Yang, Yang, Tian, and
  Chang}]{wang2024explainable}
Bo~Wang, Jing Ma, Hongzhan Lin, Zhiwei Yang, Ruichao Yang, Yuan Tian, and
  Yi~Chang. 2024.
\newblock Explainable fake news detection with large language model via defense
  among competing wisdom.
\newblock In \emph{Proceedings of the ACM on Web Conference 2024}, pages
  2452--2463.

\bibitem[{Wang and Shu(2023)}]{wang2023explainable}
Haoran Wang and Kai Shu. 2023.
\newblock Explainable claim verification via knowledge-grounded reasoning with
  large language models.
\newblock \emph{arXiv preprint arXiv:2310.05253}.

\bibitem[{Wang(2017)}]{wang2017liar}
William~Yang Wang. 2017.
\newblock \href {https://doi.org/10.18653/v1/P17-2067} {``liar, liar pants on
  fire'': A new benchmark dataset for fake news detection}.
\newblock In \emph{Proceedings of the 55th Annual Meeting of the Association
  for Computational Linguistics (Volume 2: Short Papers)}, pages 422--426,
  Vancouver, Canada. Association for Computational Linguistics.

\bibitem[{Wardle(2020)}]{wardle2020understanding}
Claire Wardle. 2020.
\newblock \emph{Understanding Information Disorder: Essential Guides}.
\newblock First Draft.

\bibitem[{{World Risk Poll}(2020)}]{web:gallup2020fake}
{World Risk Poll}. 2020.
\newblock `fake news' is the number one worry for internet users worldwide.
\newblock
  \url{https://wrp.lrfoundation.org.uk/news/fake-news-is-the-number-one-worry-for-internet-users-worldwide}.
\newblock Online; accessed 07 September 2024.

\bibitem[{Wu et~al.(2021)Wu, Rao, Sun, and He}]{wu2021evidence}
Lianwei Wu, Yuan Rao, Ling Sun, and Wangbo He. 2021.
\newblock Evidence inference networks for interpretable claim verification.
\newblock In \emph{Proceedings of the AAAI conference on artificial
  intelligence}, volume~35, pages 14058--14066.

\bibitem[{Yang et~al.(2019)Yang, Pentyala, Mohseni, Du, Yuan, Linder, Ragan,
  Ji, and Hu}]{yang2019xfake}
Fan Yang, Shiva~K Pentyala, Sina Mohseni, Mengnan Du, Hao Yuan, Rhema Linder,
  Eric~D Ragan, Shuiwang Ji, and Xia Hu. 2019.
\newblock Xfake: Explainable fake news detector with visualizations.
\newblock In \emph{The World Wide Web Conference}, pages 3600--3604.

\bibitem[{Yang et~al.(2022)Yang, Ma, Chen, Lin, Luo, and
  Chang}]{yang2022coarse}
Zhiwei Yang, Jing Ma, Hechang Chen, Hongzhan Lin, Ziyang Luo, and Yi~Chang.
  2022.
\newblock A coarse-to-fine cascaded evidence-distillation neural network for
  explainable fake news detection.
\newblock In \emph{Proceedings of the 29th International Conference on
  Computational Linguistics}, pages 2608--2621, Gyeongju, Republic of Korea.
  International Committee on Computational Linguistics.

\bibitem[{Zurko(2022)}]{zurko2022disinformation}
Mary~Ellen Zurko. 2022.
\newblock \href {https://doi.org/10.1109/MSEC.2022.3159405} {Disinformation and
  reflections from usable security}.
\newblock \emph{IEEE Security \& Privacy}, 20(3):4--7.

\end{thebibliography}

\appendix

\section{Extended Related Work on Fairness of AI-based Fake News Detectors}
\label{app:related-work}

Combating disinformation has been a widely recognized challenge in today's society. 
Agents of disinformation have learned that using genuine content and reframing it in misleading but indistinguishable ways is less likely to be detected by existing AI systems~\cite{wardle2020understanding}.
Another recent research by Chen and Shu~\cite{chen2024llmgenerated} also stress that the AI-generated misinformation, compared with human-written pieces, can be harder to detect by human and even AI-based classifiers.
Besides that, the harm of disinformation is often amplified during dissemination, where people may share it on their social networks without realizing it is false and even believing that it is helpful.

The study on disinformation involves not only the computer science community but also political, journalism, and socio-psychology researchers. 
Zurko~\cite{zurko2022disinformation} reviews the evolution of disinformation and misinformation since the 1970s and outlooks the technological development for combatting disinformation from the usable security perspective.
Mazurczyk et al.~\cite{mazurczyk2024disinformation} share their opinions on leveraging AI techniques in building a holistic solution to counter disinformation.

\section{Selection of the Backend LLM}
\label{app:model_selection}

We performed a round of preliminary studies on a smaller scale to compare the performance of different LLMs (with similar sizes). Specifically, we evaluated the performance of four LLMs, namely \textit{llama3.1:8b-instruct}, \textit{llama3:8b-instruct}, \textit{gemma:7b}, and \textit{mistral:7b}, using the LangChain RAG technique on the data collected on Aug 11. The comparison of overall performance is shown in Table~\ref{tab:model_selection}.


\begin{table}[h]
\centering
\def\arraystretch{1.01}
\setlength{\tabcolsep}{3pt}
\caption{Comparison of different choices of the the backend LLM (the best results are shown in \textbf{bold} font)\label{tab:model_selection}}
\vspace{0cm}
\small

\vspace{-6pt}
\resizebox{.48\textwidth}{!}{%
\begin{tabular}{lrrrr}
\hline
\makecell[lt]{\textbf{Models}\\(w. Knowledge\\Cutoff)} & 
\makecell[rt]{\textbf{llama3.1:8b-}\\\textbf{instruct}\\(Dec 2023)} & 
\makecell[rt]{\textbf{llama3:8b-}\\\textbf{instruct}\\(Mar 2023)} & 
\makecell[rt]{\textbf{gemma:7b}\\(Not specified, not\\later than Feb 2024)} & 
\makecell[rt]{\textbf{mistral:7b}\\(Not specified, not\\later than Sep 2024)} \\ \hline
Precision & \textbf{0.846} & 0.776 & 0.698 & 0.775 \\ \hline
Recall    & 0.941 & 0.930 & 0.842 & \textbf{0.980} \\ \hline
Accuracy  & \textbf{0.891} & 0.846 & 0.764 & 0.865 \\ \hline
F1 score  & \textbf{0.852} & 0.779 & 0.696 & 0.797 \\ \hline
\end{tabular}
}

\end{table}

We observed that the four evaluated LLMs performed similarly, and we chose the Llama 3.1 (llama3.1:8b-instruct) in the paper for its best overall performance. Besides this, the results also reveal that the detection of Manicod mainly relies on the natural language comprehension capabilities of LLMs rather than their intrinsic knowledge of recent events.

We have also attempted to perform the knowledge-grounded inference on larger models, e.g., 70b version of Llama3, and observed very limited improvement with significant computation resource usage and time sacrifice. Therefore, we adopt Llama 3.1 (8b) to balance the performance and cost of time and hardware resources. 

\section{Additional Details of our Dataset}
\label{sec:appendix-dataset}

\subsection{Prompts involved in Automatic Processing}
\label{app:appendix-dataset-prompt}

The prompt template used for generating the negation is shown below:

\begin{figure}
	\centering
	\vspace{-6pt}
	\includegraphics[width=1\linewidth]{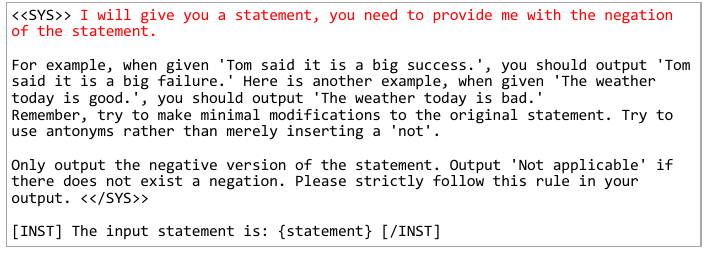}
	\vspace{-15pt}
\end{figure}

We then depict the negation generation with three real-world news headlines crawled on 1 August 2024 as follows:

\begin{figure}
	\centering
	\vspace{-9pt}
	\includegraphics[width=1\linewidth]{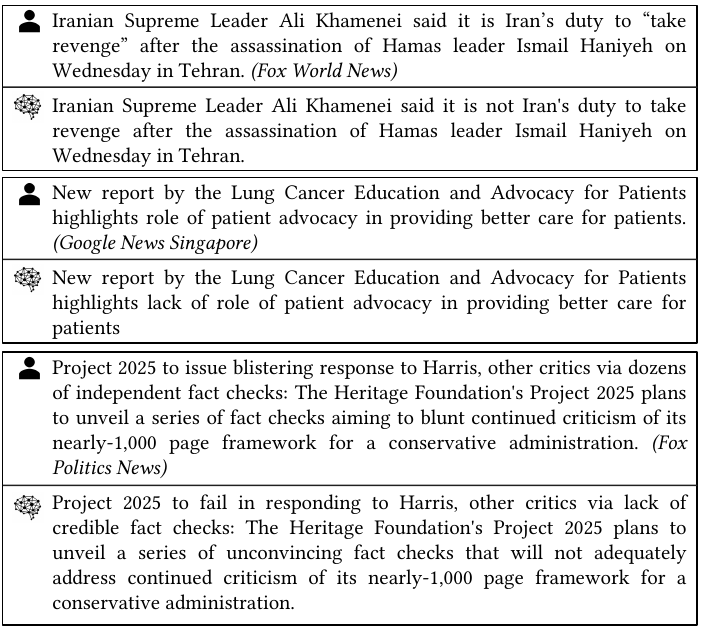}
	\vspace{-6pt}
\end{figure}

The prompt template used for automatically identifying the key context is shown below:

\begin{figure}
	\centering
	\includegraphics[width=1\linewidth]{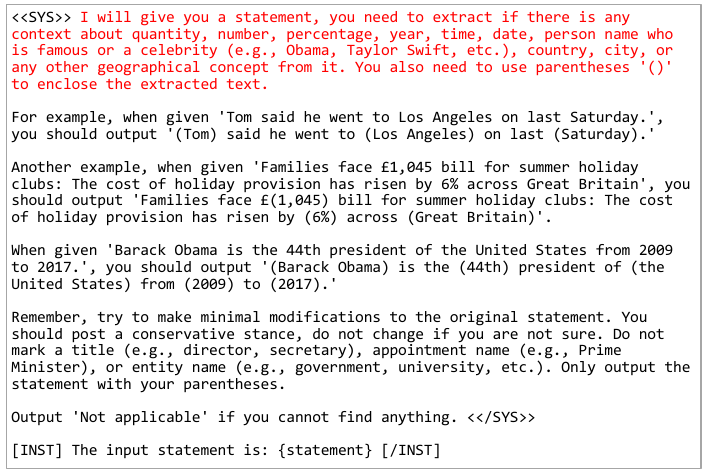}
	\vspace{-12pt}
\end{figure}

We then demonstrate the process of context alteration with three real-world news headlines collected on 25 July 2024, during which the LLM is used to automatically extract the key context and manual alteration is then involved to replace the original context with manipulated ones.

\begin{figure}
	\centering
	\vspace{-6pt}
	\includegraphics[width=0.95\linewidth]{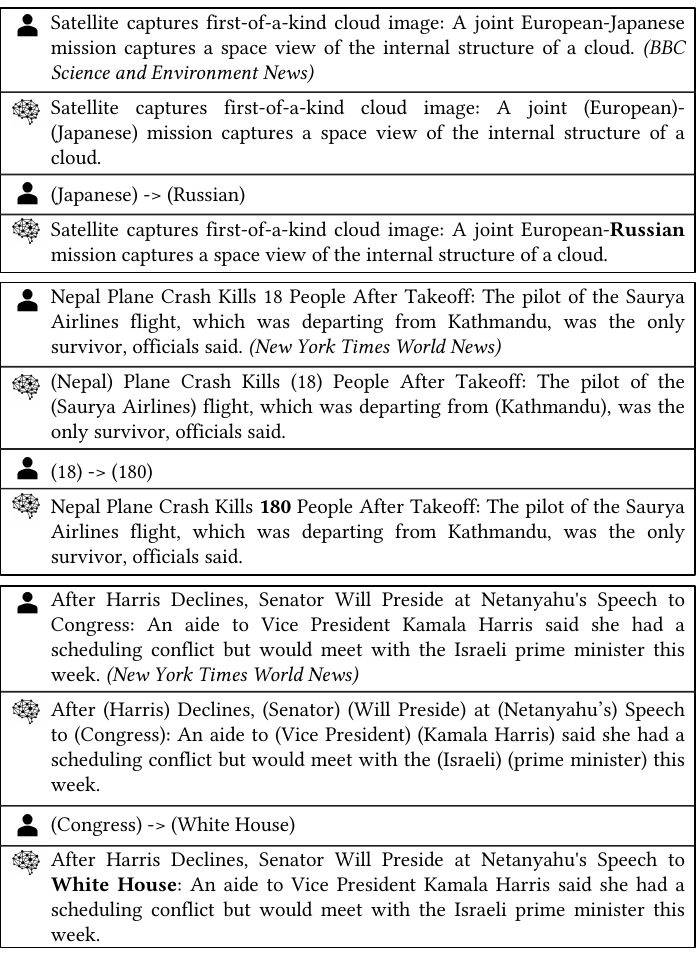}
	\vspace{-6pt}
\end{figure}

\subsection{Improper Negation Generation: An Example}
\label{app:appendix-dataset-negation-example}

We remark that the adoption of LLMs during the dataset creation is only for generating negations and identifying key context. Inappropriate or illogical negations are excluded during manual review. Below is a typical example:

\begin{figure}
	\centering
	\vspace{-6pt}
	\includegraphics[width=0.98\linewidth]{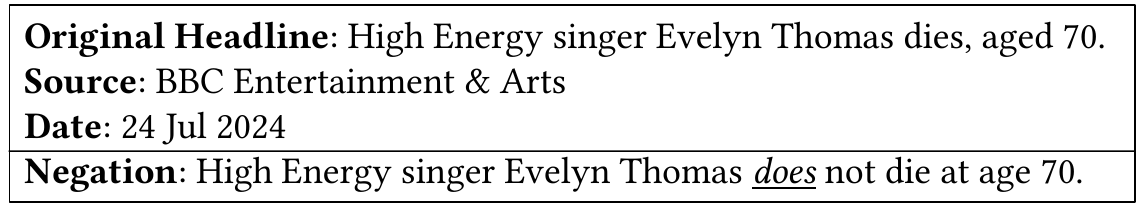}
	\vspace{-6pt}
\end{figure}

The original news headline satisfies our requirement as a claim with its veracity verifiable. However, the negation is not a reasonably manipulated real-world claim. So, that negation is not included in our dataset, although it can be correctly identified by \toolname. 

\subsection{Context Alteration and Potential Bias}
\label{app:appendix-dataset-context}

\paragraph{Manual Alteration and Rules}
Three members of our team independently performed the alteration by referring to the manipulated content observed from Google fact-check tools~\cite{google2024bigbench} and accordingly mimicking the generation of manipulated content in the real world. Each member performs iteration for 1/3 samples and reviews the remaining 2/3 of samples to reach an agreement. 

The modification is based on a set of rules defined by us. For example, for the numbers, we consider many manipulations are made to exaggerate facts. So, we increase the numbers' magnitude in manipulated samples. For entity names like countries and celebrities, our manipulations are based on personal inception about certain entities, e.g., swapping country names, swapping the name of the current US president with the historical one, etc. 

\paragraph{Potential Bias}
The context alteration, although is performed manually, can minimize the potential bias of the adopted LLMs in detecting the manipulated contents, given a reasonable assumption that LLMs may have an aptitude for identifying contents generated by themselves. 
Nonetheless, we remark that the subjective bias of our members may unavoidably be involved, which may indeed reflect the process of how manipulated contents were generated. Our work focuses on the detection of manipulated content. The bias and social aspects involved in generating manipulated content are not in our scope.

\section{Time Performance}
\label{app:appendix-time-lapsed}

Our experiments are partially performed on an Nvidia H100 GPU (80GB VRAM) and an Nvidia A100 GPU (40GB VRAM).  
As part of our evaluation, we record the time lapsed during the context construction (i.e., document chunking and vectorizing during the RAG) and the knowledge-grounded inference. 
Given the top three search results to be crawled for retrieval, the median value of time spent on context construction is \avgragtime on the tested H100 GPU server, with the 25\% and 75\% percentiles of 5.774 seconds and 27.0 seconds, respectively. The vectorization of some retrieved contents takes a longer time mainly because the contents on those webpages are too complicated and contain too many irrelevant contents like advertisements. We did not observe an obvious difference in the time performance of context construction between the tested H100 GPU and the A100 GPU.
In contrast to the context construction, we observed a much more uniform distribution of the time elapsed for the LLM's knowledge-grounded inference. The median of inference time recorded is \avginferencetime on the H100 server and 3.58 seconds on the A100 server.

\section{Case Study of Inference Anomaly}
\label{sec:appendix-case-study}
\paragraph{Failure in Reasoning Chronological Order \& Confusion of Premise and Hypothesis}
We find a case failing due to this reason in our evaluation. The claim and the inference output are presented as follows.

\begin{figure}
	\centering
	\includegraphics[width=0.98\linewidth]{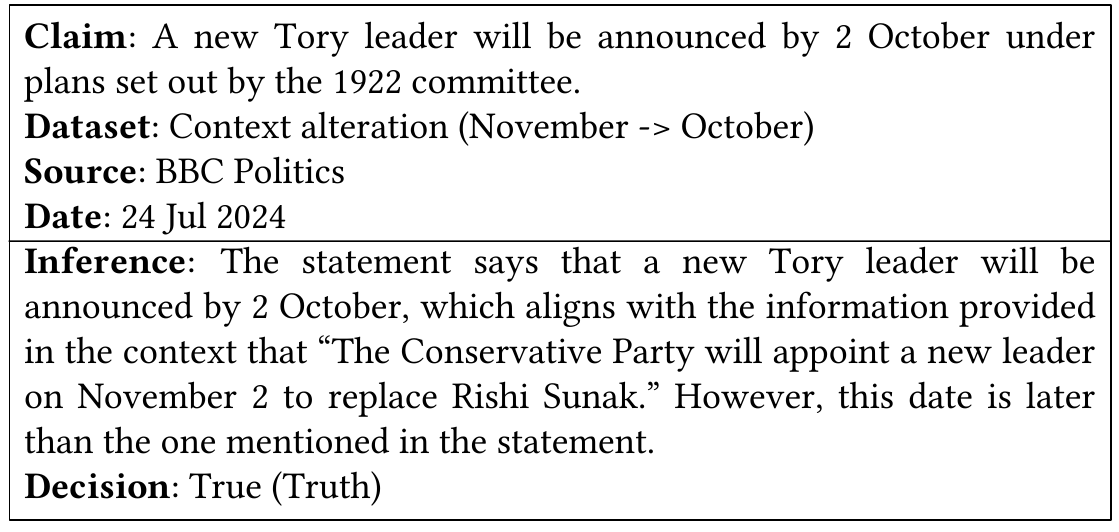}
	\vspace{-6pt}
\end{figure}

Given the original news headline states that ``\textit{the leadership of the Tory will be announced by 2 November},'' we altered the due date to one month earlier, making the claim a piece of disinformation because the deadline provided in the claim contradicts the ground truth. However, by interpreting the inference output, we find that the adopted LLM failed to reason the chronological order of these two dates, especially the logical relationship when these two dates are used as deadlines, resulting in a false prediction. Besides that, from the last sentence of the inference, we observed that the LLM became confused about the premise and hypothesis. It mistakenly treated the input claim as the premise (i.e., the ground truth), and attempted to determine the veracity of the retrieved online knowledge (i.e., the context supplied to the LLM).

\paragraph{Failure in Comprehending Synonyms}
Although LLMs are known to be powerful in natural language comprehension, we still find some cases failed because of mistakes in understanding synonyms. We present an example below.

\begin{figure}
	\centering
	\vspace{-6pt}
	\includegraphics[width=0.98\linewidth]{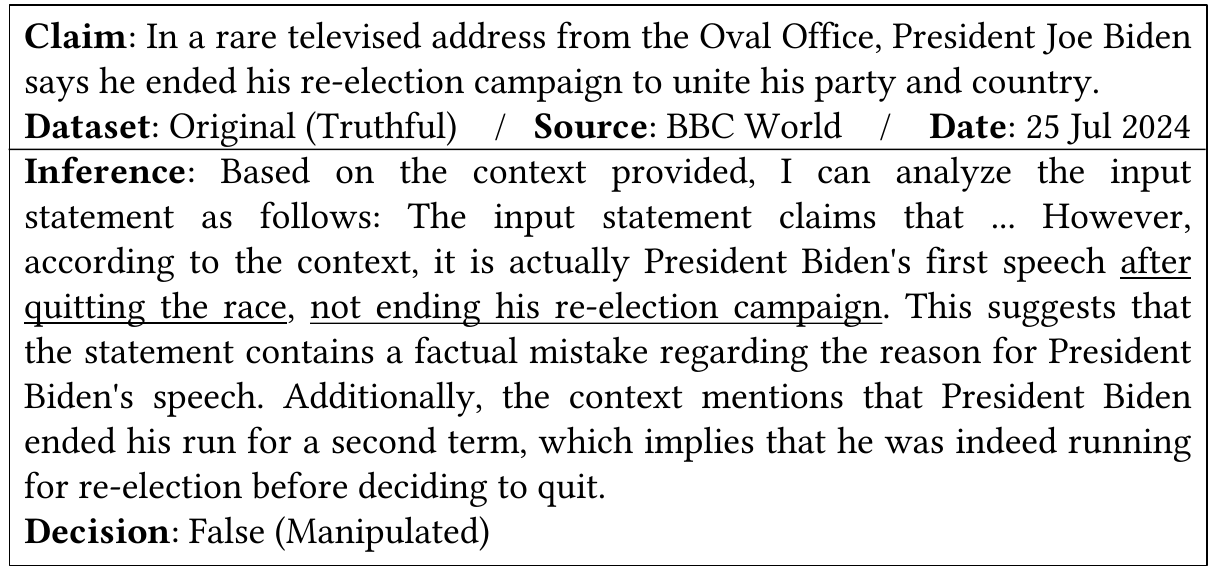}
	\vspace{-6pt}
\end{figure}

From the inference output, we find the LLM correctly reasoned the claim against the retrieved knowledge. However, it failed to understand the meaning of the term ``ending the re-election'', and mistakenly differed it from ``quitting the race'' for the presidential election although these two terms express the exactly same meaning. As a result, the LLM determined it as false information.

\paragraph{Failure in Reasoning in Numbers}
LLMs sometimes may not be able to compare the values of multiple numbers. Below we show an example from the COVID-Scientific dataset observed during the baseline benchmarking.

\begin{figure}
	\centering
	\vspace{-6pt}
	\includegraphics[width=0.98\linewidth]{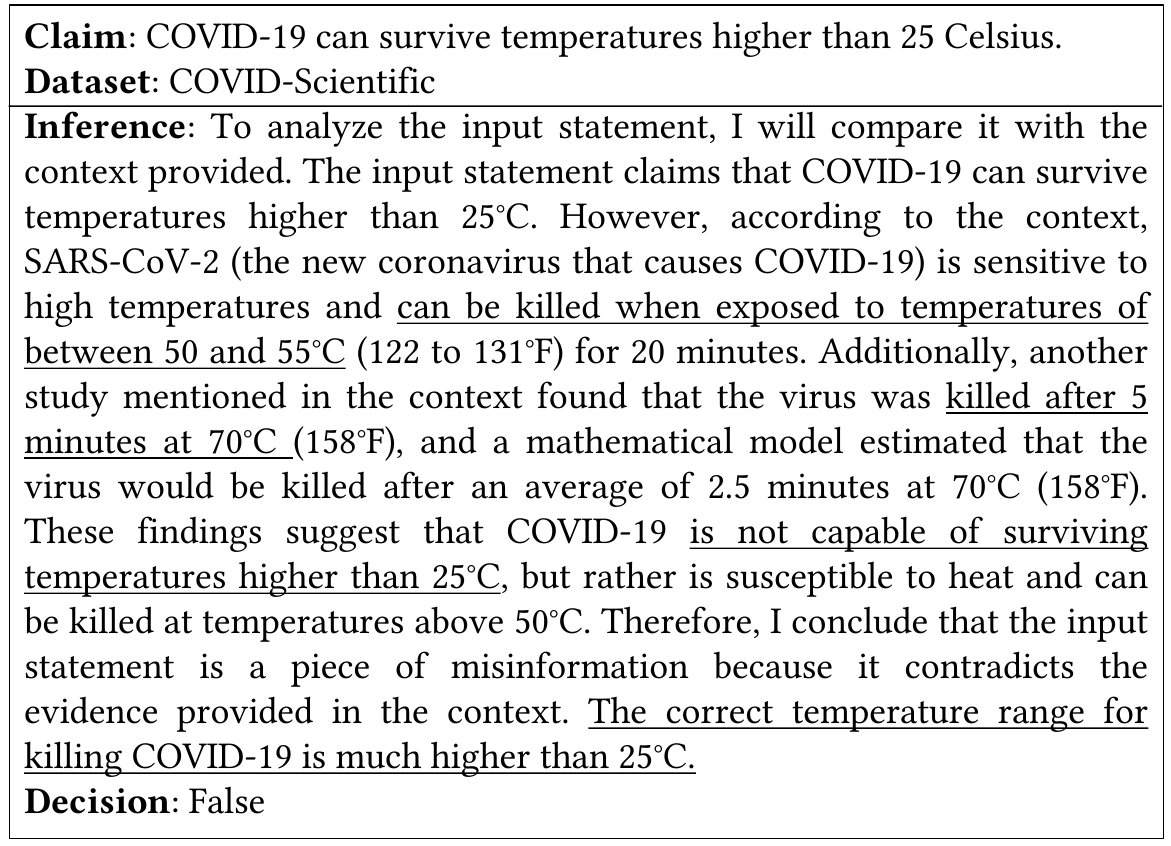}
	\vspace{-6pt}
\end{figure}

In the inference, we found that \toolname managed to find suitable knowledge from the Internet but failed to compare the values of 25\textcelsius~and 50\textcelsius. Based on the knowledge in the context, the lowest temperature that may kill the virus is 50\textcelsius, which makes the claim indisputably true. However, the adopted LLM did not manage to compare these two values and mistakenly determined the claim as a false information.

\section{Confusion Matrix of the Evaluation (RQ1)}
\label{app:confusion_matrix}

We count the occurrence of true positive, true negative, false positive, and false negative cases based on the evaluation of our proposed dataset containing \allclaims claims, i.e., \alltruenews recent real-world news headlines and derived \allfake manipulated fake news. The confusion matrix is shown in the table below:


\begin{table}[h]
\centering
\def\arraystretch{1.01}
\setlength{\tabcolsep}{3.4pt}
\vspace{0cm}
\small

\begin{tabular}{lrr}
\hline
 & \textbf{Positive (fake)} & \textbf{Negative (truth)} \\ \hline
\textbf{True} & 3956 & 1483 \\ \hline
\textbf{False} & 1017 & 314 \\ \hline
\end{tabular}

\end{table}

Using those results, we calculate the precision, recall, and F1 score of \toolname and present them in~\autoref{sec:results}.


\end{document}